# A Novel Hybrid Feature Importance and Feature Interaction Detection Framework for Predictive Optimization in Industry 4.0 Applications

Zhipeng Ma
SDU Center for Energy Informatics, the
Maersk Mc-Kinney Moller Institute
University of Southern Denmark
Odense, Denmark
zhma@mmmi.sdu.dk

Bo Nørregaard Jørgensen
SDU Center for Energy Informatics, the
Maersk Mc-Kinney Moller Institute
University of Southern Denmark
Odense, Denmark
bnj@mmmi.sdu.dk

Zheng Grace Ma
SDU Center for Energy Informatics, the
Maersk Mc-Kinney Moller Institute
University of Southern Denmark
Odense, Denmark
zma@mmmi.sdu.dk

*Abstract*—Advanced machine learning algorithms are increasingly utilized to provide data-based prediction and decision-making support in Industry 4.0. However, the prediction accuracy achieved by the existing models is insufficient to warrant practical implementation in real-world applications. This is because not all features present in real-world datasets possess a direct relevance to the predictive analysis being conducted. Consequently, the careful incorporation of select features has the potential to yield a substantial positive impact on the outcome. To address the research gap, this paper proposes a novel hybrid framework that combines the feature importance detector - local interpretable model-agnostic explanations (LIME) and the feature interaction detector - neural interaction detection (NID), to improve prediction accuracy. By applying the proposed framework, unnecessary features can be eliminated, and interactions are encoded to generate a more conducive dataset for predictive purposes. Subsequently, the proposed model is deployed to refine the prediction of electricity consumption in foundry processing. The experimental outcomes reveal an augmentation of up to 9.56% in the $R^2$ score, and a diminution of up to 24.05% in the root mean square error.

## I. INTRODUCTION

The utilization of process digitalization in Industry 4.0 requires continuous data acquisition and evaluation. With advanced data analysis methods, data from production flows can be processed to provide data-based support for decision-making processes. To ensure the success of data analyses, comprehensive data pre-processing is essential.

However, as the volume of available production data rapidly increases, the number of features has to be reduced beforehand since not all of them are relevant to the analysis and some may even negatively impact predictions. Therefore, reducing the number of features through a reasonable feature selection procedure is crucial to enhance the effectiveness of the data analysis process. One of the primary objectives of feature selection is to identify the important variables that affect prediction accuracy [1]. Additionally, feature interaction detection is another critical task because the integration of features may produce better results than the behavior of one or all of them by capturing nonlinear relationships between the input and target variables. New features can be created based on the interactions to enhance prediction accuracy [2]. Moreover, these explanations can also assist engineers in comprehending which factors are beneficial for improving aspects of industrial processing, such as production quality and energy consumption, to optimize the production processes.

In recent years, several algorithms have been developed for detecting feature importance and feature interactions. Regarding feature importance detection, it is aimed at choosing a subset of variables from the input that can efficiently describe the input data while minimizing the effects of noise or irrelevant variables and still producing accurate predictions [3]. A framework for benchmarking attribution methods (BAM) has been proposed in [4], comprising a crafted dataset and models trained with known relative feature importance, and three complementary metrics to quantitatively evaluate attribution approaches. S. Hooker et al. [5] have evaluated interpretability methods by assessing how the accuracy of a retrained model degrades as features estimated to be important are removed, and the fraction of the most important pixels are replaced with a fixed uninformative value. Furthermore, the local interpretable model-agnostic explanation (LIME) [6] is a local surrogate model trained to explain the predictions of classifiers or regressors, and it can be globally extended based on a submodular pick.

Regarding feature interaction detection, the interaction between features refers to a group of features that have a joint non-additive impact on predicting an outcome [7]. The anchor explanation [8] is a model-agnostic feature interaction detector that uses an if-then rule that sufficiently "anchors" the prediction locally. A Bayesian personalized feature interaction selection (BP-FIS) mechanism [9] has been proposed based on Bayesian variable selection theory. In addition, a neural interaction detector [10] has been developed to extract the interactions by directly computing the learned weights of a feedforward multilayer neural network.

Furthermore, the developed algorithms regarding feature importance and feature interactions have been utilized in industrial sectors. Depth-based isolation forest has been developed in [11] to analyze the important features in the refrigerator manufacturing process. A distributed and parallel feature extraction algorithm is proposed to detect the important time series in industrial processes [12]. A hybrid graph and rule-based approach [13] has been developed to identify the interaction features from the CAD part models.

Although feature importance and feature interaction detection methods are popular research fields in interpretable machine learning, few studies have investigated how to combine them in a feature selection pipeline. Additionally, the technology of applying both feature selection and feature interaction procedures to optimize the prediction models and analyze the significant factors impacting industrial processing is lacking.

To fill the research gap mentioned above, this article proposes a hybrid framework to combine LIME and NID algorithms to improve prediction accuracy. The framework optimizes the acquired dataset by removing unimportant features and encoding interactions. Subsequently, LIME is applied again to select significant variables for prediction. Finally, this framework is applied to optimize the prediction of electricity consumptions in the foundry processing, resulting in an improvement of up to 9.56% in the $R^2$ score and a reduction of up to 24.05% in the root mean square error.

The remainder of the article is organized as follows. Section II introduces the principles of LIME and NID in detail. The proposed framework and components are elaborated and demonstrated in section III. Section IV explains the experiments based on a foundry processing dataset. The results and discussion are presented in section V. Section VI concludes the work and recommendations for future work.

## II. RELATED WORK

### A. Feature Importance Detection – LIME

The local interpretable model-agnostic explanation (LIME) [6] is a surrogate explanation algorithm designed to explain the predictions of any classifier or regressor by learning an interpretable model in the vicinity of the predicted instance. It can also be expanded to provide global interpretations by employing submodular optimization to select a representative set of data instances.

The explanation produced by LIME is obtained by:

$$\xi(x) = \underset{g \in G}{argmin}\, L(f, g, \pi_x) + \Omega(g) \qquad (1)$$

where $g$ represents a model to be explained, $G$ is a class of potentially linear interpretable models, $L$ refers to the weighted square loss, $f$ is the denotation of g, $\pi_x$ represents the exponential smoothing kernel and the weight of the features, and $\Omega(g)$ denotes the measure of complexity.

The objective of the LIME is to minimize the loss function $\xi(x)$. To achieve this, the data instance being explained, $x$, is firstly transformed into an interpretable form. Afterward, samples are drawn around $x$ uniformly at random to approximate the loss $L$. Subsequently, lasso regression is applied to train the perturbed samples via least squares. The importance of each feature is determined by the absolute values of the weights, with the sign indicating whether the influence is positive or negative (a positive sign indicates a positive effect).

Explaining feature importance based on a single data instance is inadequate for global evaluation. To address this limitation, Submodular Pick (SP) is proposed to extend the application of LIME for the global explanation. SP picks a diverse, representative set of non-redundant explanations to showcase how the model behaves globally, which applies greedy optimization [14] to pick instances that cover the important components, avoiding selecting instances with similar explanations.

LIME is a highly effective algorithm for generating explanations. One of its primary strengths is its model-agnostic nature, which allows users to substitute the prediction algorithm without affecting the local, interpretable model employed for the explanation. The flexibility of the algorithm enables it to be used across a wide range of applications. Additionally, LIME produces concise and contrastive results, which facilitates user comprehension and makes the explanations more user-friendly.

### B. Feature Interaction Detection – NID

The neural interaction detector (NID) [10] explains feature interactions by directly interpreting the learned weights of a feedforward multilayer neural network. Feature interactions are created at hidden units with nonlinear activation functions, i.e., ReLu, and propagated layer by layer to the output. The interaction strength $\omega(j)$ is defined to quantify the interactions.

$$\omega_i(j) = z_i^{(1)} \mu(|\mathbf{W}_{i,j}^{(1)}|) \qquad (2)$$

$$\mu(|\mathbf{W}_{i,j}^{(1)}|) = \min(|\mathbf{W}_{i,j}^{(1)}|) \qquad (3)$$

$$z^{(l)} = |\mathbf{w}|^T \cdot |\mathbf{W}^{(L)}| \cdot |\mathbf{W}^{(L-1)}| \cdot |\mathbf{W}^{(l+1)}| \qquad (4)$$

where $\omega_i(j)$ represents the interaction strength of a potential interaction $j$ at the $i$-th unit in the first hidden layer, $\mu(\cdot) = min(\cdot)$ refers to the minimizing function, $z_i^{(l)}$ is the aggregated weight representing the influence of a hidden unit $i$ at the $l$-th hidden layer, $\mathbf{w}$ represents the coefficients for the final output, and $\mathbf{W}$ refers to the weight matrix.

A ranking of interaction candidates is generated through a greedy algorithm. It only considers the top-ranked interactions of every order based on their interaction strengths at each hidden unit to reduce the search space of potential interactions. Furthermore, the selection of the top-K interactions is carried out through the implementation of a cut-off procedure by analyzing interaction strengths.

The NID algorithm can detect both pairwise and higher-order interactions without requiring complex model training, which represents a significant advantage over other algorithms. Additionally, NID's processing speed is noteworthy, as it outperforms many competing algorithms while producing similar top-ranking interactions.

## III. METHODOLOGY

### A. General Pipeline

This paper proposes a novel hybrid feature importance and feature interaction detection framework, combining LIME and NID to improve the prediction accuracy. In order to mitigate the negative impact of unimportant features, those with low importance scores are removed based on LIME results. Additionally, to take advantage of interactive information, interaction features are grouped and embedded into the dataset using NID. Furthermore, the framework reuses LIME to extract the important features to further improve the prediction performance. The general pipeline is illustrated in Fig.1.

In the reconstruction stage, for a dataset with n features and a machine learning (ML) prediction model, the global form of the LIME algorithm is applied to detect feature importance by analyzing data instances and trained prediction models. Meanwhile, the NIP method is utilized to extract feature interactions. Next, the top-$k$ important features and top-$m$ interactions are selected to construct dataset II. The interactive features are created based on Cartesian products [15, 16]. For instance, the interaction set $I_i = \{F_1, F_2, ..., F_t\}$ involves $t$ different features with $F_l$ ($l = 1, 2, ..., t$) representing the set of values for the $l$-th feature. The values of the integrated feature $F_{I_i}$ are calculated as $F_1 \odot F_2 \odot ... \odot F_t$, where $\odot$ denotes the element-wise multiplication operation.

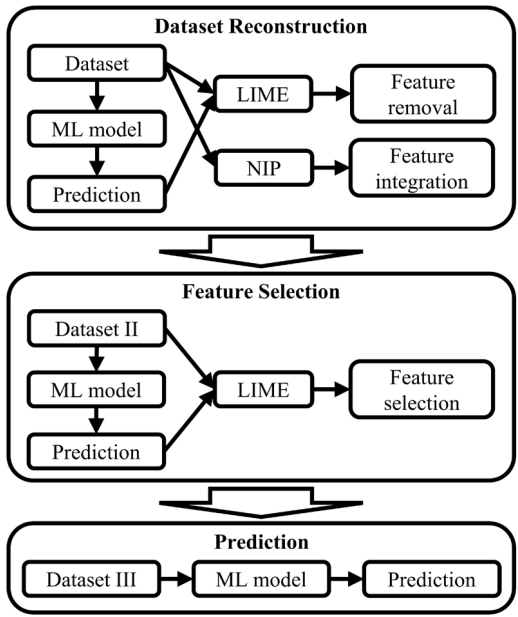

Fig. 1. Pipeline of the proposed framework.

Moving on to the feature selection stage, LIME is applied again to rank the feature importance in dataset II. Subsequently, to optimize the prediction performance, the number of unimportant features to remove is then determined through comparison. When the threshold of removed features is set as $k'$, $t$ ($t = 0, 1, ..., k'$) least important features from LIME results are removed separately for prediction, and the optimized dataset (dataset III) is selected based on the set of removed features that yields the highest prediction accuracy.

### B. Parameter Setting Suggestions

In practical applications, the thresholds used to select important and interaction features can be set flexibly based on input data and prediction models. After conducting a number of experiments, several suggestions have been made regarding how to set these thresholds. Firstly, in the data reconstruction stage, it is recommended to remove approximately 10% of the least important features and embed all detected interaction features into the dataset. Removing too many features may result in the loss of important information from the original dataset. Furthermore, the NID algorithm produces relatively few interaction sets in the experiments because of the cutoff procedure, so all integrated interaction features can be embedded in the dataset. Secondly, it is recommended that the threshold for removed features, denoted as $k'$, in the feature selection stage is set to a value smaller than half of the number of features in dataset II, which keeps a balance between runtime and optimization performance based on the experimental experience.

## IV. EXPERIMENTS

### A. Dataset Description

The dataset used in this paper is obtained from the casting process operations of one of Northern Europe's largest foundries. The objective of the experiments is to predict the electricity consumption in the production flows, which is framed as a regression task. The input features are presented in TABLE I, whereas the output features are displayed in TABLE II. The dataset consists of 18 parameters related to the casting process provided by the industrial partner as inputs, and the outputs representing the energy consumption are voltage and current. The dataset comprises 57,601 data instances.

TABLE I. INPUT FEATURES IN THE DATASET

| Features | Units |
|---|---|
| TemperatureAct | °C |
| FurnaceWeight | kg |
| FurnaceIsolationResistance | MΩ |
| FurnaceQuantity 1-8 | ton |
| FurnaceTemperature 1-7 | °C |

TABLE II. OUTPUT FEATURES IN THE DATASET

| Features | Units |
|---|---|
| FurnaceVoltage | V |
| FurnaceCurrent | A |

### B. Data Processing Settings

The programming environment in the experiment is Python (version 3.9.13). 43,353 data instances are used as training data and the remaining 14,248 instances are used as test data. Training LIME model should use the trained prediction model as input. To compare the stability and generalization of the LIME algorithm and the entire optimization framework, three different regression algorithms, including AdaBoost, random forest regression and decision tree regression, are used in the experiments. 'FurnaceVoltage' and 'FurnaceCurrent' are predicted separately with the same parameter settings.

The hyper-parameters in the experiments are set as follows. The number of estimators in AdaBoost and the random forest is both set as 100. In the LIME algorithm, the mode is set as 'regression', the method of submodular pick is sampling, and the sample size is set as 1,000. The threshold for feature removal in the feature selection stage is set to 10 and 12 for predicting voltage and current respectively based on the suggestions in section III.B.

## V. RESULTS AND DISCUSSION

### A. Results

TABLE III and TABLE IV represent the feature importance ranking based on the LIME algorithm for voltage and current prediction respectively. A low weight shows the low importance of the features in the corresponding prediction model. Therefore, as section III.B (Parameter Setting Suggestions) introduces, the feature with the lowest weight (No. 1) is removed from the dataset.

TABLE V and TABLE VII present the feature interaction sets and their interaction strengths using the NID algorithm for voltage and current prediction respectively. 8 interaction sets are produced in voltage predictions and 4 sets are detected in current predictions. Each interaction set is generated to a new feature following the procedure in section III.A (General Pipeline). Then all the generated features are embedded to the dataset whose least important feature has been removed.

TABLE III. FEATURE IMPORTANCE RESULTS FOR VOLTAGE PREDICTION

| No. | AdaBoost | | Random Forest | | Decision Tree | |
|---|---|---|---|---|---|---|
| | *Features* | *Weights* | *Features* | *Weights* | *Features* | *Weights* |
| 1 | FurnaceQuantity6 | 19.96 | FurnaceTemperature1 | 17.93 | FurnaceTemperature4 | 25.40 |
| 2 | FurnaceTemperature6 | 21.06 | FurnaceTemperature3 | 19.14 | FurnaceQuantity6 | 26.06 |
| 3 | FurnaceTemperature5 | 21.91 | FurnaceQuantity6 | 19.18 | FurnaceTemperature6 | 33.72 |
| 4 | FurnaceWeight | 22.28 | FurnaceTemperature4 | 20.21 | FurnaceWeight | 34.06 |
| 5 | FurnaceTemperature7 | 23.29 | FurnaceWeight | 24.61 | FurnaceQuantity1 | 36.75 |
| 6 | FurnaceTemperature1 | 23.61 | FurnaceTemperature7 | 25.24 | FurnaceTemperature3 | 38.59 |
| 7 | FurnaceQuantity5 | 25.24 | FurnaceTemperature6 | 29.26 | FurnaceTemperature7 | 41.73 |
| 8 | FurnaceTemperature4 | 28.47 | FurnaceQuantity5 | 34.39 | FurnaceQuantity5 | 44.36 |
| 9 | FurnaceTemperature3 | 29.94 | FurnaceTemperature5 | 36.43 | FurnaceTemperature5 | 48.04 |
| 10 | FurnaceQuantity3 | 34.73 | FurnaceTemperature2 | 41.26 | FurnaceTemperature1 | 53.26 |
| 11 | FurnaceQuantity8 | 49.20 | FurnaceQuantity1 | 48.63 | FurnaceQuantity4 | 62.80 |
| 12 | FurnaceQuantity1 | 52.09 | FurnaceQuantity8 | 49.99 | FurnaceTemperature2 | 77.49 |
| 13 | TemperatureAct | 70.61 | FurnaceIsolationResistance | 50.01 | FurnaceQuantity8 | 87.28 |
| 14 | FurnaceIsolationResistance | 91.58 | FurnaceQuantity3 | 55.89 | FurnaceIsolationResistance | 156.63 |
| 15 | FurnaceQuantity4 | 115.00 | FurnaceQuantity4 | 59.00 | TemperatureAct | 157.57 |
| 16 | FurnaceTemperature2 | 155.54 | TemperatureAct | 94.59 | FurnaceQuantity3 | 204.20 |
| 17 | FurnaceQuantity7 | 231.28 | FurnaceQuantity7 | 254.68 | FurnaceQuantity7 | 319.42 |
| 18 | FurnaceQuantity2 | 1148.25 | FurnaceQuantity2 | 975.72 | FurnaceQuantity2 | 958.76 |

TABLE IV. FEATURE IMPORTANCE RESULTS FOR CURRENT PREDICTION

| No. | AdaBoost | | Random Forest | | Decision Tree | |
|---|---|---|---|---|---|---|
| | *Features* | *Weights* | *Features* | *Weights* | *Features* | *Weights* |
| 1 | FurnaceQuantity1 | 11.73 | FurnaceQuantity6 | 16.79 | FurnaceTemperature2 | 29.06 |
| 2 | FurnaceTemperature5 | 12.08 | FurnaceTemperature4 | 17.48 | FurnaceTemperature3 | 29.79 |
| 3 | FurnaceTemperature7 | 12.47 | FurnaceTemperature5 | 17.89 | FurnaceQuantity6 | 30.56 |
| 4 | FurnaceTemperature1 | 13.20 | FurnaceTemperature2 | 18.34 | FurnaceQuantity3 | 30.68 |
| 5 | FurnaceQuantity3 | 16.51 | FurnaceTemperature3 | 18.36 | FurnaceTemperature1 | 43.22 |
| 6 | FurnaceTemperature4 | 17.57 | FurnaceTemperature1 | 19.30 | FurnaceTemperature5 | 43.38 |
| 7 | FurnaceTemperature6 | 20.76 | FurnaceTemperature7 | 21.40 | FurnaceTemperature4 | 43.85 |
| 8 | FurnaceTemperature3 | 25.67 | FurnaceQuantity1 | 32.55 | FurnaceTemperature7 | 52.83 |
| 9 | FurnaceQuantity6 | 29.70 | FurnaceTemperature6 | 33.57 | FurnaceQuantity1 | 91.70 |
| 10 | FurnaceQuantity8 | 59.13 | FurnaceQuantity3 | 49.16 | FurnaceWeight | 94.08 |
| 11 | FurnaceWeight | 72.29 | FurnaceQuantity4 | 49.27 | FurnaceQuantity4 | 95.42 |
| 12 | TemparatureAct | 76.56 | FurnaceQuantity8 | 57.47 | FurnaceQuantity8 | 127.24 |
| 13 | FurnaceTemperature2 | 82.25 | TemperatureAct | 126.44 | TemperatureAct | 131.17 |
| 14 | FurnaceQuantity4 | 88.31 | FurnaceWeight | 127.23 | FurnaceTemperature6 | 172.08 |
| 15 | FurnaceQuantity7 | 120.47 | FurnaceQuantity5 | 163.50 | FurnaceQuantity8 | 229.93 |
| 16 | FurnaceQuantity5 | 149.71 | FurnaceIsolationResistance | 192.62 | FurnaceIsolationResistance | 263.23 |
| 17 | FurnaceIsolationResistance | 153.60 | FurnaceQuantity7 | 320.40 | FurnaceQuantity7 | 290.54 |
| 18 | FurnaceQuantity2 | 482.19 | FurnaceQuantity2 | 593.45 | FurnaceQuantity2 | 502.98 |

In Fig. 2-5, 'AB' represents AdaBoost, 'RF' refers to the random forest, and 'DT' denotes the decision tree. Fig.2 and Fig.4 demonstrate the trend of $R^2$ score with increasing the number of removed features in the feature selection procedure, while Fig.3 and Fig.5 show the trend of root mean square error (RMSE). The line charts illustrate that the $R^2$ score tends to increase and the RMSE tends to decrease as more unimportant features are removed. To get the trade-off of both accuracy and precision, the case optimizing both $R^2$ score and RMSE in the iterations is selected as the optimized prediction.

Fig.6 and Fig.7 compare the optimized prediction performance with the original results. The blue bars represent the original $R^2$ scores under distinct prediction models and the orange bars represent the optimized results. The error bars denote the standard deviation of multiple experimental results. In the category labels, 'V' refers to voltage prediction and 'C' refers to current prediction. 'AB', 'RF' and 'DT' denote

AdaBoost, random forest and decision tree respectively. Fig.6 illustrates that $R^2$ scores are improved in all experiments and Fig.7 displays the reduction of RMSE.

TABLE V. VOLTAGE PREDICTION FEATURE INTERACTION RESULTS

| No. | Interaction sets | Interaction Strengths |
|---|---|---|
| 1 | {FurnaceIsolationResistance; FurnaceTemperature 1, 2, 3, 5, 6} | 1927.53 |
| 2 | {TemperatureAct; FurnaceIsolationResistance; FurnaceTemperature 1, 2, 3, 5, 6} | 560.70 |
| 3 | {FurnaceWeight, FurnaceIsolationResistance} | 289.65 |
| 4 | {TemperatureAct; FurnaceIsolationResistance; FurnaceTemperature 1, 2, 3, 5, 6} | 275.49 |
| 5 | {TemperatureAct; FurnaceIsolationResistance; FurnaceTemperature 1, 2, 3, 4, 5, 6} | 109.93 |
| 6 | {TemperatureAct, FurnaceIsolationResistance, FurnaceTemperature 1, 2, 3, 4, 5, 6, 7} | 107.91 |
| 7 | {FurnaceWeight, FurnaceIsolationResistance, FurnaceQuantity8} | 106.81 |
| 8 | {TemperatureAct; FurnaceWeight; FurnaceIsolationResistance; FurnaceQuantity 1, 2, 3, 4, 5, 6, 7, 8; FurnaceTemperature 1, 2, 3, 4, 5, 6, 7} | 100.23 |

TABLE VI. CURRENT PREDICTION FEATURE INTERACTION RESULTS

| No. | Interaction sets | Interaction Strengths |
|---|---|---|
| 1 | {FurnaceIsolationResistance, FurnaceTemperature2} | 24.96 |
| 2 | {FurnaceIsolationResistance; FurnaceTemperature 1, 2, 3, 5, 6} | 24.13 |
| 3 | {FurnaceIsolationResistance, FurnaceTemperature4} | 14.53 |
| 4 | {TemperatureAct; FurnaceWeight; FurnaceIsolationResistance; FurnaceQuantity 1, 2, 3, 4, 5, 6, 7, 8; FurnaceTemperature 1, 2, 3, 4, 5, 6, 7} | 13.62 |

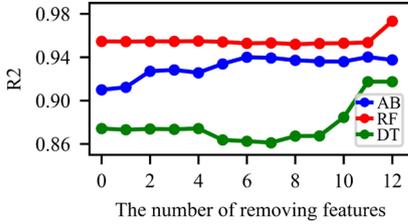
Fig. 2. $R^2$ scores on voltage prediction for different removing features.

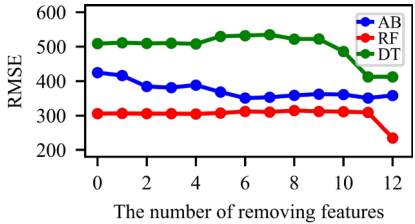
Fig. 3. RMSE on voltage prediction for different removing features.

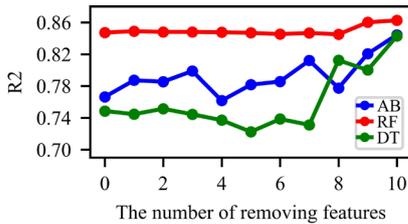
Fig. 4. $R^2$ scores on current prediction for different removing features.

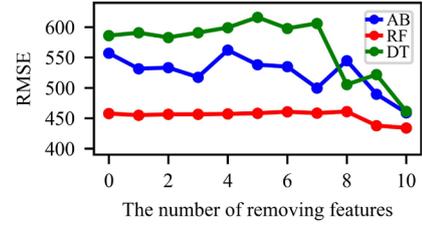
Fig. 5. RMSE on current prediction for different removing features.

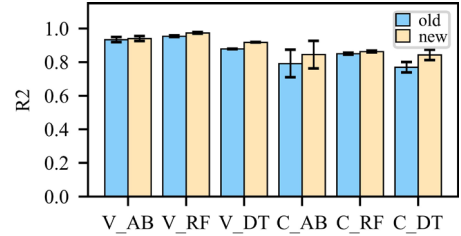
Fig. 6. Comparison of R2 scores in the experiments.

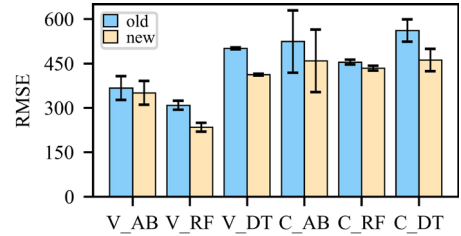
Fig. 7. Comparison of RMSE in the experiments.

TABLE VII shows the numerical performance improvements in experiments and the number of removed features. The $R^2$ scores are reduced at 0.67%-9.56%, and the RMSE are improved at 4.50%-24.05%, which indicates the optimization performance of the proposed framework.

TABLE VII. ACCURACY IMPROVEMENTS AND FEATURE DELETION IN THE EXPERIMENTS

| Experiments | $R^2$ score | RMSE | No. feature deletion |
|---|---|---|---|
| V_AB | 0.67% | 4.53% | 6 |
| V_RF | 2.06% | 24.05% | 12 |
| V_DT | 4.49% | 17.74% | 11 |
| C_AB | 6.67% | 12.42% | 10 |
| C_RF | 1.56% | 4.50% | 10 |
| C_DT | 9.56% | 17.80% | 10 |

*B. Discussion*

The LIME algorithm analyzes both the dataset and the trained prediction model. TABLE III and TABLE IV show that this method generates distinct but similar results when applied to the same dataset using different prediction models. For instance, the most important feature in voltage prediction is 'FurnaceQuantity2' in all three prediction models, and their importance weight is much larger than the second important ones. However, their least important features are ranked differently because the different model mechanism and prediction outcomes results in the different surrogate models in LIME. Such difference does not significantly affect the framework performance, because the optimization pipeline relies on the prediction model as well, which represents that the feature processing is unique for each prediction algorithm.

The NID algorithm processes the dataset through a neural network framework, independently of the trained prediction models, resulting in unique feature interaction results for the same dataset. All interaction features are embedded in the dataset because the interaction sets are much less than the features in the input dataset based on TABLE V and TABLE VI. This process provides rich information without significantly increasing prediction runtime. Furthermore, since the majority of interaction strengths are similar, it is not practical to classify the interaction sets as important or unimportant. Therefore, eliminating the interaction set with the lowest strength is also not a viable solution.

Fig.2 to Fig.5 display the trend of $R^2$ scores and RMSE in the feature selection procedure. The $R^2$ scores are gradually improved while the RMSE is decreased as more unimportant features are detected by LIME in the second phase, indicating that most of the removed variables impact the prediction negatively. Moreover, the embedded interaction features may introduce noises into the model since the multiplication operation for feature integration amplifies the noise. Hence, reusing the LIME for feature selection is necessary and effective to improve the prediction quality. By combining LIME and NID algorithms, the proposed Hybrid feature importance and feature interaction detection framework improves the $R^2$ scores and reduces the RMSE in all six experiments as Fig.6 and Fig.7 illustrate, which shows the stability and reliability of the proposed framework model.

The proposed framework is capable of optimizing both the $R^2$ score and RMSE. The R2 score measures the goodness of fit of a regression model, while the RMSE represents the accuracy of predictions. If the goal is to improve the fitting quality, it is important to select the feature set that improves the R2 score. On the other hand, if the emphasis is on improving the accuracy of predictions, it is important to select the feature set that reduces the RMSE.

In addition to prediction optimization, the LIME algorithm also provides interpretable explanations of the dataset, elucidating the primary variables that influence energy consumption. As evidenced by TABLE III and TABLE IV, 'FurnaceQuantity2' emerges as the predominant factor affecting energy consumption. Consequently, it is recommended that the corresponding operations are optimized to enhance overall electricity efficiency.

VI. CONCLUSION

This paper proposes a novel hybrid framework combining the feature importance and feature interaction detectors to optimize the prediction performance through refined feature reconstruction. It is applied to optimize voltage and current predictions in casting processing flows. By removing unimportant features based on LIME results, their negative impact on predictions is mitigated. Additionally, incorporating interaction features derived from NID results contributes more valuable information for prediction purposes. Finally, the feature selection process leveraging the LIME algorithm further optimizes prediction performance. Experimental results demonstrate robust optimization performance, with $R^2$ score reductions ranging from 0.67% to 9.56% and RMSE improvements between 4.50% and 24.05%.

The proposed framework serves not only as an optimization pipeline but also an explanatory tool. First, by eliminating and embedding features, it enhances prediction performance, facilitating improved strategic planning. Second, the generated feature importance rankings and interaction sets elucidate the relationships between variables, enabling industrial stakeholders to better comprehend the operational mechanisms in production flows and make more informed decisions in various industrial sectors.

In future work, the proposed framework could be further enhanced by optimizing the feature selection process using intelligent and automated algorithms. Moreover, an optimized approach for generating interaction features may yield additional insights into variable relationships, further enriching the information available for predictions.

ACKNOWLEDGMENT

This paper is part of the project "Data-driven best-practice for energy-efficient operation of industrial processes - A system integration approach to reduce the CO2 emissions of industrial processes" (Case no.64020-2108) by the Danish funding agency, the Energy Technology Development and Demonstration (EUPD) program, Denmark.

REFERENCES

[1] M. R. Islam, A. A. Lima, S. C. Das, M. F. Mridha, A. R. Prodeep, and Y. Watanobe, "A Comprehensive Survey on the Process, Methods, Evaluation, and Challenges of Feature Selection," IEEE Access, vol. 10, pp. 99595-99632, 2022.

[2] M. Calder, M. Kolberg, E. H. Magill, and S. Reiff-Marganiec, "Feature interaction: a critical review and considered forecast," Comput Netw, vol. 41, no. 1, pp. 115-141, Jan 15 2003.

[3] S. Solorio-Fernandez, J. A. Carrasco-Ochoa, and J. F. Martinez-Trinidad, "A survey on feature selection methods for mixed data," Artif Intell Rev, vol. 55, no. 4, pp. 2821-2846, Apr 2022.

[4] M. Yang and B. Kim, "Benchmarking attribution methods with relative feature importance," arXiv preprint arXiv:1907.09701, 2019.

[5] S. Hooker, D. Erhan, P.-J. Kindermans, and B. Kim, "A benchmark for interpretability methods in deep neural networks," Advances in neural information processing systems, vol. 32, 2019.

[6] M. T. Ribeiro, S. Singh, and C. Guestrin, ""Why Should I Trust You?" Explaining the Predictions of Any Classifier," in Kdd'16: Proceedings of the 22nd Acm Sigkdd International Conference on Knowledge Discovery and Data Mining, pp. 1135-1144, 2016.

[7] C. Molnar, Interpretable Machine Learning: A Guide for Making Black Box Models Explainable (2nd ed.), 2022. [Online]. Available: https://christophm.github.io/interpretable-ml-book

[8] M. T. Ribeiro, S. Singh, and C. Guestrin, "Anchors: High-Precision Model-Agnostic Explanations," Aaai Conf Artif Inte, pp. 1527-1535, 2018.

[9] Y. F. Chen, P. J. Ren, Y. Wang, and M. de Rijke, "Bayesian Personalized Feature Interaction Selection for Factorization Machines," Proceedings of the 42nd International Acm Sigir Conference on Research and Development in Information Retrieval (Sigir '19), pp. 665-674, 2019.

[10] M. Tsang, D. Cheng, and Y. Liu, "Detecting statistical interactions from neural network weights," arXiv preprint arXiv:1705.04977, 2017.

[11] M. Carletti, C. Masiero, A. Beghi, and G. A. Susto, "Explainable Machine Learning in Industry 4.0: Evaluating Feature Importance in Anomaly Detection to Enable Root Cause Analysis," Ieee Sys Man Cybern, pp. 21-26, 2019.

[12] M. Christ, A. W. Kempa-Liehr, and M. Feindt, "Distributed and parallel time series feature extraction for industrial big data applications," arXiv preprint arXiv:1610.07717, 2016.

[13] V. B. Sunil, R. Agarwal, and S. S. Pande, "An approach to recognize interacting features from B-Rep CAD models of prismatic machined parts using a hybrid (graph and rule based) technique," Comput Ind, vol. 61, no. 7, pp. 686-701, Sep 2010.

[14] U. Feige, "A threshold of ln n for approximating set cover," Journal of the ACM (JACM), vol. 45, no. 4, pp. 634-652, 1998.

[15] R. Wang, B. Fu, G. Fu, and M. Wang, "Deep & cross network for ad click predictions," in Proceedings of the ADKDD'17, 2017, pp. 1-7.

[16] Y. F. Luo et al., "AutoCross: Automatic Feature Crossing for Tabular Data in Real-World Applications," in Kdd'19: Proceedings of the 25th Acm Sigkdd International Conferencce on Knowledge Discovery and Data Mining, pp. 1936-1945, 2019.